\pdfoutput=1

\documentclass[11pt]{article}

\usepackage{EMNLP2022}

\usepackage{times}
\usepackage{latexsym}

\usepackage[T1]{fontenc}

\usepackage[utf8]{inputenc}

\usepackage{microtype}

\usepackage{inconsolata}
\usepackage{amsmath}
\usepackage{adjustbox}
\usepackage{tabularx}
\usepackage{makecell}
\usepackage{multirow}
\usepackage{enumitem}
\usepackage{caption}
\usepackage{bbding}
\usepackage{color}
\usepackage{booktabs}
\usepackage{listings}
\usepackage{comment}

\title {Distinguish Sense from Nonsense: \\Out-of-Scope Detection for Virtual Assistants}

\author{Cheng Qian$\dagger$, Haode Qi$\dagger$, Gengyu Wang$\dagger$ Ladislav Kunc$\dagger$, Saloni Potdar$\ddagger$ \\
         $\dagger$IBM Watson, $\ddagger$Apple Inc. \\ 
         {\texttt{\{cheng.qian, haode.qi, gengyu, lada\}@ibm.com}, \texttt{s\_potdar@apple.com}} }

\begin{document}
\maketitle
\begin{abstract}
Out of Scope (OOS) detection in Conversational AI solutions enables a chatbot to handle a conversation gracefully when it is unable to make sense of the end-user query. Accurately tagging a query as out-of-domain is particularly hard in scenarios when the chatbot is not equipped to handle a topic which has semantic overlap with an existing topic it is trained on.
We propose a simple yet effective OOS detection method that outperforms standard OOS detection methods in a real-world deployment of virtual assistants.
We discuss the various design and deployment considerations for a cloud platform solution to train virtual assistants and deploy them at scale.
Additionally, we propose a collection of datasets that replicates real-world scenarios and show comprehensive results in various settings using both offline and online evaluation metrics.

\end{abstract}

\section{Introduction}

In the context of task-oriented dialog, Out of Scope (OOS) detection is the problem of identifying end-user queries that are beyond the scope of a chatbot. While this problem is generally studied under the umbrella of ``out of domain'' detection in machine learning, we show that unique challenges arise in real-world applications. 
We study this problem in the context of our enterprise virtual assistant (VA) platform which is used by 10,000+ customers to design chatbots. In this setting, the natural language understanding models comprising of In Scope (IS) and OOS detection modules, need to determine whether an input query belongs to a set of pre-defined intents or if it is out of scope for the chatbot.

Real-world success of OOS systems often involves measuring how good they are at \textit{containment}, i.e., the user queries are resolved and contained by the chatbot while minimizing human interventions. Since containment rate can be only observed after launching the VA online, offline metrics such as IS accuracy and OOS accuracy are needed while designing and developing the models. 


 The average designer using an enterprise VA platform is not a machine learning expert. This leads to a variety of challenges in the provided user data, which constitutes the need for robust algorithms.
Firstly, end-users often provide data that is heavily imbalanced or noisy for both IS and OOS detection. 

While designing VA for enterprise use-cases, IS and OOS examples often naturally belong to the same domain. Such OOS samples are called In Domain OOS (ID-OOS) as opposed to Out-of-Domain OOS (OOD-OOS) which are relatively easier OOS samples from a different domain (\citet{zhang-etal-2022-pre-trained}). Designers expect the VA to detect these relevant but unsupported topics (ID-OOS) even though it has high semantic overlap with IS examples. Finally, while entities defined by the designer play an important role for a real-world VA, they are often ignored in academic OOS settings. We show that entities must be modeled in conjunction with IS and OOS classification.

In this paper, we discuss the challenges of designing a real-world OOS detection system in depth and common approaches taken to design such a system. 
 We propose a simple but effective algorithmic modification for OOS detection in a real-world deployed system. This system models entities, intent and OOS classification jointly and addresses the challenges around data. We propose a comprehensive benchmark based on public datasets and show that our method outperforms standard approaches while being simple to deploy and maintain.


\section{Challenges}
\subsection{Metrics}
\textbf{Containment and Disambiguation}
For businesses, the key performance index (KPI) metric is typically different from the common machine learning metrics used to test the performance of the algorithm. Businesses use containment rate to measure chatbot performance - the portion of conversations not handed off or escalated to a human agent for quality reasons.
Among offline evaluation metrics, IS performance provides the best estimate for containment rate. 
Disambiguation is a mechanism to increase containment by asking end-users clarification questions and providing more than one relevant intent.
This has to be counter-balanced with high OOS performance so that we don't provide a set of predictions in the form of IS classes for an OOS query. This is essential to appear "intelligent" and handle conversations gracefully.

\par \noindent \textbf{In Domain Out of Scope Detection} refers to detecting OOS samples with high semantic overlap with IS examples in the same domain (ID-OOS). ID-OOS queries are often harder to detect than the easier Out of Domain OOS (OD-OOS) samples. The algorithm should be able to identify ID-OOS and also generalize well to unseen OOD-OOS.

\subsection{Data Considerations}
The average designer of an enterprise VA platform doesn't need to have ML background, hence the expectations of labeled data are very different from an ML expert end-user. 

\par \noindent \textbf{Class Imbalance} is often extreme in data provided by VA designer, with some intent classes having more number of examples than others.

\par \noindent \textbf{Data-scarce scenarios} Labeling data is often expensive for enterprises who desire good performance with very few labelled examples per class, and often no OOS labeled data. 

\par \noindent \textbf{Noisy data} Unlike public datasets, enterprise datasets have semantically similar intents due to overlap in business use cases. Additionally, real-world end-user input queries to VAs usually contain spelling errors, intentionally repeated characters, emojis, and slang. Proper normalization is required to improve robustness of OOS  algorithms.

\subsection{Computational Efficiency}
While developing the OOS detection algorithm for an enterprise VA platform, we need to strike a good trade-off between cost of serving the model and performance of the model. Based on our experience, VA platforms are expected to handle training sets of more than 10k training examples and more than 1000 classes.

\par \noindent \textbf{Model size \& memory}: 
There are over 100,000 customer-specific models deployed in production and each chatbot serves millions of queries per month. Hence low maintenance, training and inference costs can increase profitability.  

\par \noindent \textbf{Training time:} Designers typically make changes in an iterative fashion, designing the VA through trial and error. For an interactive experience in the product, the OOS detection component needs to train in ~1 minute (\citet{DBLP:journals/corr/abs-2012-03929}). 

\par \noindent \textbf{Inference time:} During the inference, each query passes through all the natural language understanding (NLU) components - IS classification, OOS detection, entity recognition and spellchecking, and needs to provide the predictions in 10 milliseconds.  
 
 
\subsection{Entities and OOS Detection}
Entities are designed to represent nouns from end-user inputs and are crucial for VAs to respond accordingly and haven't been studied extensively in OOS detection.

\par \noindent \textbf{Terminologies} Designers can define entities with special terminologies that are out of vocabulary of any other public or private corpus. This requires OOS detection methods to differentiate such terminologies from gibberish sentences. 

\par \noindent \textbf{Synonyms} 
The OOS detection algorithm is expected to produce similar detection scores across the multitudes of synonyms of the same entity. 

\par \noindent \textbf{Numeric Values} System entities like date, number, time etc. are pre-configured in a VA to cover a wide range of concepts. 
However, there is no one-size-fits-all solutions for system entities. E.g., the system entity "11" can be a part of domain specific terminology "operating system Windows 11". The OOS detection algorithm needs to be aware of these system entity values and decide the relevance of the sentence based on the context. 

We introduce several potential solutions for handling entities in OOS detection and analyze their advantages and disadvantages.

\par \noindent \textbf{Concatenation of all entity synonyms} 
In the context of Binary OOS detection, we add one synthetic IS example in to the training data by concatenating all entity synonyms provided for a chatbot. Context independent features such as uni-grams, bi-grams and mean/max pooling of word-embeddings will help recognizing these entities as IS at the runtime. This simple approach works well empirically but has the disadvantage of ignoring the context and semantic meaning.



\par \noindent \textbf{Synonyms and Entity proxies in intent templates} \label{fixed_synonyms} 
In enterprise VA, an entity can be defined with multiple synonyms.
In our product, we support entity proxies, which is a definition of a certain entity and its associated synonyms that are considered equal. This greatly simplifies training data definition at the cost of potential instability at runtime: our intent detection and OOS algorithm should return the same confidence and same predicted label if one synonym is replaced by another. 
For the example in Table \ref{tab:entity ood example}, if "cell phone" is defined as an entity proxy, VA designer only references the symbol "cell phone" in training examples, and at runtime "i want an iphone 11" gets the exact same prediction as "i want an iphone xr".

 
\section{OOS Detection Algorithms}
 \label{sec-oos-alg}
OOS detection algorithms can be broadly classified into single-stage and multi-stage.
\vspace{-2.5mm}
\subsection{Single-stage OOS} All the IS classes and optionally the OOS class are used together train a single model to determine if the incoming query belongs to one of the IS intents or is OOS.




\par \noindent \textbf{Multiclass Classification} In this approach, the algorithm treats the OOS examples as an additional class as explored in (\citet{DBLP:journals/corr/abs-2106-08616}, \citet{DBLP:journals/corr/abs-2105-05601}), alongside the IS classes
to train a multi-class classification model for both IS intent detection and OOS detection.\footnote{\url{ https://docs.microsoft.com/en-us/azure/cognitive-services/luis/concepts/intents\#none-intent}} 
This approach trains a single algorithm for both OOS detection and IS detection tasks. In practice, this approach is susceptible to over-fitting to the provided OOS examples and might not generalize well to unseen OOS queries. Additionally, it can fail in the presence of severe class-imbalance.

\par \noindent \textbf{In-scope Classification utilizing output distribution} This type of methods trains a classifier on IS data which outputs a probability vector with low maximum probability or high entropy for an OOS input, as explored in \citet{10.1007/978-1-4471-2099-5_1}, \citet{DBLP:journals/corr/HendrycksG16c}, \citet{lee2018training}, \citet{DBLP:conf/naacl/YilmazT22}. 
These methods train a single model 
for IS detection and applies a threshold on output probability distribution statistics (such as max and entropy) for OOS detection. However, in practice, training data typically has semantically overlapped intents 
which will mislead the system and increase unnecessary human agent intervention as shown in Table \ref{tab:ambiguous examples}. 

\subsection{Multi-stage OOS} Multi-stage OOS method uses a binary classifier to determine if a query is IS or OOS in the first stage. In the subsequent stages we determine which of the IS intent is the closest match.

\par \noindent \textbf{Binary Classification (IS/OOS)}: A binary classifier is trained using the IS examples and OOS examples as explored in \citet{DBLP:conf/esann/TaxD99}. The classification result is used to determine if end-user query is OOS or ID.
In case there are no OOS training examples, the binary OOS classifier can be replaced with an one-class classifier or other unsupervised methods. Another solution for the lack of OOS training data is synthetic OOS training examples, refer to Section \ref{limit} for more discussion. 

\par \noindent \textbf{In-scope Classification plus unsupervised methods on internal (hidden state) representation} trains a classifier based on IS training examples, and utilizes internal representation (for example, concatenation of hidden states from several layers of a neural network) of the IS classifier for an unsupervised OOS detection algorithm, like autoencoder with reconstruction loss, distance based approach \cite{wu-etal-2022-revisit}, \cite{DBLP:journals/corr/abs-2106-14464}, and density based approach \cite{DBLP:journals/corr/abs-1906-00434}.





\subsection{Our Approach}

We show a simple modification to the multi-stage OOS to improve the performance of the system and alleviate problems with the other approaches mentioned previously. Our approach \textbf{Binary Classification(In Scope/OOS) discounting on In Scope scores} treats OOS classification as a binary classification problem like the previous formulation. However, the binary classification score of the OOS detection algorithm is used to discount the IS classification score to determine the final IS score (more details to follow). In case no training OOS examples are available our OOS detection algorithm becomes one-class classification. This formulation is related to calibration \cite{DBLP:journals/corr/abs-2006-09462} that trains a new model to reject inputs when the model is over-confident. However, our approach applies the OOS output as discounting factor instead of binary score leading to better performance in the context of enterprise VA as shown in Table \ref{exp_perf}.

In terms of the OOS classification component, we implement a distance based approach based on sentence embedding of both IS and OOS training examples (if labeled). At training time, we first apply the trick described in Section \ref{fixed_synonyms} to preprocess entities among other text normalization steps, then query the sentence embeddings from a sentence encoder. For each IS example, we store the linear combination of its sentence embedding and the mean embedding of its corresponding intent class in an approximate nearest neighbor(ANN) search index. If there are OOS training examples, we store their sentence embeddings in the same ANN search index. At runtime, a query is preprocessed the same way as training examples, the cosine distance to the nearest neighbor will be used as OOS score to discount the output from the IS classifier. If the nearest neighbor is OOS, we add an additional constant to the corresponding nearest distance, to discount the confidence more. The discounting step uses the OOS score \(cos\_dist\) and IS classifier output confidence \(\textbf{conf}\), we apply the formula below to produce the final output confidence vector \(\textbf{final\_conf}\) as follows:

\vspace{-0.1in}
\begin{equation}
\footnotesize{
\textbf{final\_conf} = (1 - f (max(cos\_dist, 0) )) \cdot \textbf{conf}
}
\end{equation}
\vspace{-0.3in}
\begin{equation}
\footnotesize{
f(x) = 
\begin{cases}
x & x \geq 0.5\\
sigmoid(a \cdot (x-0.5)) &\text{otherwise}
\end{cases}
}
\label{discountingformula}
\end{equation}
, where \(a\) is a constant that can be tuned for different applications. 
The motivation for Formula \ref{discountingformula} is to reduce the amount of discounting on IS confidence (comparing to a linear discounting function), when OOS classifier predicts a low cosine distance (thus high similarity) for an utterance.


Typically, a fixed threshold T on the final output confidences is used in real world applications to determine whether an input utterance is predicted as IS or OOS: a new input is deemed OOS if its final output confidence is less than T. Theoretically, this threshold is not critical to machine learning metrics, especially threshold independent metrics. Even for threshold dependent metrics, this threshold can always be tuned in accordance with the scale of final output confidence to achieve the same results. However in practice, as a commercial VA platform, a fixed threshold reduces the maintenance cost of a chatbot and only a small fraction of the chatbot designers will try to tune the threshold. In our product, 0.2 threshold is set as the default value.

\begin{table}[]
\centering
\begin{adjustbox}{max width=0.45\textwidth}
\begin{tabular}{ll}
\toprule
Query                                             & Intent               \\ \midrule
\makecell[l]{I need assistance with my \\retirement account }           & retirement account \\ 
\makecell[l]{I need to talk to a agent about \\ my retirement account }       & agent            \\ 
\bottomrule
\end{tabular}
\end{adjustbox}
\caption{
The two queries shown are semantically overlapped.
For the query "I need to talk to a assistant about my retirement account", the correct intent should be "agent" but one can expect "retirement account" and "agent" having similar probability. For approaches that rely on probability vector to detect OOS input, these examples can mislead them to treat valid IS queries as OOS.}
\label{tab:ambiguous examples}
\end{table}

\subsection{Benchmark Dataset}



We collect 8 intent classification datasets to comprehensively evaluate the methods mentioned above regarding the challenges, including IS classification, OOS detection, and scalability. The 8 datasets include \textbf{ATIS} \cite{hemphill1990atis}, \textbf{BANKING77} \cite{Casanueva2020}, \textbf{CLINC150} \cite{larson2019evaluation}, \textbf{StackOverflow}\footnote{\url{ https://storage.googleapis.com/download.tensorflow.org/data/stack_overflow_16k.tar.gz}}, \textbf{SNIPS} \cite{DBLP:journals/corr/abs-1805-10190}, \textbf{HAR} \cite{liu2019benchmarking}, \textbf{ROSTD} \cite{gangal2020likelihood}, and \textbf{HINT3} \cite{arora2020hint3}. A summary of dataset statistics after preprocessing is provided in Table \ref{table_stat}.

To evaluate methods' performance on ID-OOS detection, we ensure all datasets contain ID-OOS examples. For datasets that only contain IS examples, we randomly choose a number of IS intents and treat them as OOS so that the number of examples in these intents are about 25\% of the whole training dataset.
The full list of chosen intents for each dataset are listed in Appendix \ref{sec:appendix_dataset}. 

We reorganize some of the datasets as follows. CLINC150 includes 2 domains, banking and credit card, we evaluate them separately along with the full set. For StackOverflow, 10\% examples in training set is stratified splited as dev set. For HAR, we remove examples with missing 'answer', and stratified split remaining examples into train, dev, and test set with a 80, 10, and 10 percentage. Different from other selected datasets, ROSTD contains 4,000 OOS examples. ROSTD-coarse is the version that only keep higher hierarchical intent types. Examples in ``reminder'' intent type from original ROSTD-coarse are treated as ID-OOS. HINT3 consists of 3 domains, including SOFMattress, Curekart and Powerplay11, so we evaluate them separately. 10\% of training examples in each of HINT3 datasets is stratified split as dev set. 
\begin{table*}[]
\begin{adjustbox}{max width=1\textwidth}
\begin{tabular}{lccc|ccc|ccc}
\toprule
Dataset                       & \multicolumn{3}{c}{Train} & \multicolumn{3}{c}{Dev}  & \multicolumn{3}{c}{Test} \\
                              & IS   & ID-OOS & OOD-OOS & IS  & ID-OOS & OOD-OOS & IS  & ID-OOS & OOD-OOS \\
                              \midrule
CLINC150-FULL                    & 11300 & 3700    & 100     & 2260 & 740     & 100     & 3390 & 1110    & 1000    \\
CLINC150-BANKING                 & 400   & 100     & 0       & 400  & 100     & 600     & 400  & 100     & 1350    \\
CLINC150-CREDIT                  & 400   & 100     & 0       & 400  & 100     & 600     & 400  & 100     & 1350    \\
ATIS                          & 4053  & 425     & 0       & 458  & 42      & 0       & 812  & 81      & 0       \\
BANKING77                     & 6533  & 2089    & 0       & 1160 & 380     & 0       & 2320 & 760     & 0       \\
Stack Overflow                & 5400  & 1800    & 0       & 600  & 200     & 0       & 6000 & 2000    & 0       \\
SNIPS                         & 9371  & 3713    & 0       & 500  & 200     & 0       & 500  & 200     & 0       \\
ROSTD                         & 23621 & 6900    & 0       & 3238 & 943     & 1500    & 6661 & 1960    & 3090    \\
ROSTD-coarse                  & 23621 & 6900    & 0       & 3238 & 943     & 1500    & 6661 & 1960    & 3090    \\
HAR                           & 15893 & 4592    & 0       & 1986 & 575     & 0       & 1985 & 576     & 0       \\
HINT3 (SOFMattress)           & 229   & 66      & 0       & 26   & 7       & 0       & 158  & 73      & 166     \\
HINT3 (Powerplay11)           & 317   & 102     & 0       & 38   & 14      & 0       & 244  & 31      & 708     \\
HINT3 (Curekart)              & 415   & 125     & 0       & 45   & 15      & 0       & 390  & 62      & 539     \\
\bottomrule
\end{tabular}
\end{adjustbox}
\caption{\textbf{Dataset Statistics.} We preprocess all datasets (details in~\ref{sec:appendix_dataset}) and numbers reflect their sizes.}
\label{table_stat}
\end{table*}

\subsection{Evaluation metrics}
\label{subsec-eval-metrics}
Based on current literature, there are 2 types of commonly used metrics for OOS detection.

\textbf{Threshold dependent metrics} are metrics calculated with predicted labels e.g. accuracy. These metrics compare the probability score against a threshold to determine whether a query is considered OOS or not. Also threshold dependent metrics encourage joint evaluation of intent detection and OOS detection that are more suitable under the context of VA. Following the literature (\citet{wu-etal-2022-revisit}, \citet{zhou-etal-2022-knn}, \citet{DBLP:journals/corr/abs-2105-14289}), the threshold dependent metrics are listed here:

\textbf{Overall Accuracy} is the percentage of examples being correctly classified. For an IS query, it's predicted correctly if and only if the predicted IS label is correct and the query is predicted as IS. For an OOS query, it should be predicted as OOS to make a correct prediction. \textbf{IS Accuracy} is the percentage of correctly predicted IS examples out of all IS examples. 
\textbf{IS F1} is the macro averaged F1 scores across all IS intents.
\textbf{OOS F1} is the F1 score for OOS examples.

\textbf{Threshold independent metrics} are metrics calculated with a vector of scores each of which measures how confident or likely an OOS algorithm considers a query irrelevant. Such a score is often a number between 0 and 1 where 1 represents IS and 0 represents OOS. This paper follows the literature (\citet{DBLP:journals/corr/abs-2106-14464}, \citet{ryu-etal-2018-domain}, \citet{DBLP:journals/corr/LiangLS17}, \citet{NEURIPS2018_abdeb6f5}) and defines IS as the positive class and OOS as the negative class. We use the metrics for evaluating OOS detection performance:
\textbf{FPRN}, where N is an integer between 0 and 100, is the false positive rate(FPR) when the true positive rate(TPR) is at least N\%. A false positive is an OOS example predicted as IND. We use FPR90 and FPR95.
\textbf{AUROC} is the area under the Receive Operating Characteristic curve, which measures TPR against FPR at different thresholds.
\textbf{AUPR\_IN} and \textbf{AUPR\_OUT} are metrics measuring area under the precision-recall curve, when IS and OOS are considered as the positive class, respectively.

\begin{table}[!t]
\centering
\begin{adjustbox}{max width=0.45\textwidth}
\begin{tabular}{l}
\toprule
        \textit{Training Example} \\ Can I buy a cell phone ? \\ \midrule
        \textit{Training Entities} \\   \makecell[l]{entity: cell phone\\
        synomyms: iphone, samsung, galaxy, iphone XR, iphone 11, etc..
        } \\ \midrule
        \textit{Inference Queries} \\ \makecell[l]{
                A galaxy is a huge collection of gas, dust, stars and their solar systems.\\
        what is the latest model of galaxy s series?\\
        }
        \\ 
        \bottomrule
\end{tabular}
\end{adjustbox}
\caption{
VA designer defines the entity "cell phone" with synonyms. 
The 1st query contains the word "galaxy" but it is OOD-OOS. The 2nd with "galaxy" is ID-OOS. 
} 
\label{tab:entity ood example}
\end{table}

\begin{table*}[]
\begin{adjustbox}{max width=1\textwidth}
\begin{tabular}{lcccccccccc}
\toprule
Method & Overall Acc. & IS Acc. & IS F1 & OOS F1 & OOS recall & FPR90 & FPR95 & AUROC & AUPR\_IN & AUPR\_OUT  \\ 
      \midrule
        Binary & 82.45 & 86.92 & 77.87 & 76.84 & 74.18 & 22.76 & 30.29 & 91.70 & 91.21 & 88.16 \\      
        Multiclass & 74.34 & 90.36 & 72.81 & 74.08 & 64.84 & - & - & - & - & - \\ 
        IS clf + Max & 79.17 & 85.93 & 77.99 & 70.02 & 65.15 & 32.66 & 44.45 & 87.53 & 87.15 & 80.27 \\ 
        \bf{Discounting (Our Approach)} & 84.70 & 86.43 & 79.71 & 80.63 & 79.31 & 20.07 & 27.20 & 92.64 & 91.07 & 89.90  \\ 
\bottomrule
\end{tabular}
\end{adjustbox}
\caption{\textbf{Performance on all datasets} This table compares the discounting method against binary classification, multiclass classification, the IS classifier + max confidence on the full test sets.
}
\label{exp_perf}
\end{table*}

\begin{table*}[]
\begin{adjustbox}{max width=1\textwidth}
\begin{tabular}{lcccccccccc}
\toprule
Method & Overall Acc. & IS Acc. & IS F1 & OOS F1 & OOS recall & FPR90 & FPR95 & AUROC & AUPR\_IN & AUPR\_OUT  \\ 
      \midrule
        Binary & 53.45 & 72.70 & 47.97 & 45.26 & 38.40 & 59.24 & 78.80 & 81.37 & 72.07 & 86.98  \\ 
        Multiclass & 52.97 & 74.65 & 49.44 & 56.15 & 41.15 & ~ & ~ & ~ & ~ &   \\ 
        IS clf + Max & 54.36 & 75.13 & 53.46 & 57.90 & 43.66 & 56.38 & 73.74 & 76.97 & 60.33 & 86.01  \\ 
        \bf{Discounting (Our Approach)} & 61.46 & 69.38 & 51.79 & 60.05 & 54.94 & 50.84 & 71.02 & 82.83 & 69.32 & 88.97 \\ 
\bottomrule
\end{tabular}
\end{adjustbox}
\caption{\textbf{Performance on HINT3} This table compares the discounting method against the Multiclass classification method, the binary classification method, the IS classifier + max confidence on the full test sets.
}
\label{exp_perf_hint3}
\end{table*}






\subsection{Experiments and Results}
\label{results}

We conduct experiment on the benchmark datasets to compare different OOS problem formulations listed in Section \ref{sec-oos-alg} (Our discounting approach, Binary Classification (IS/OOS), Multiclass Classification\footnote{Threshold independent metrics for Multiclass classification is omitted, as our IS classifier outputs confidence vectors (which do not sum up to 1) instead of predicted probability, thus it involves no such component as OOS scores.}
, and IS Classification utilizing output distribution). As a comparison for the OOS problem formulation only, we keep the IS classification algorithm and OOS algorithm same as our production setup across the 4 formulations, without focusing on the exact choice or implementation of the IS and OOS algorithms.  For the discounting method, we use our production intent detection and OOS detection as is. For multi-class classification method, we consider OOS examples as an additional IS intent. For the IS Classification utilizing output distribution formulation, we train an IS classifier with IS training examples and take the max of output confidence vector as the OOS score. 

\par \noindent \textbf{Offline Evaluation} \label{offline_eval}
Table \ref{exp_perf} reports the simple average of metrics across all our benchmark datasets. The discounting approach achieves better performance across most metrics. We report the average metrics in Table \ref{exp_perf_hint3} on the subset of the 3 HINT3 datasets, which are designed to represent real world imbalanced datasets.

Table \ref{exp_discounting_vs_kp1} compares the Multi-Class formulation against our formulation on all datasets. Our approach performs on par on ID-OOS but generalize better to OOD-OOS. In real world application, limited ID-OOS is provided by customer during training but the algorithm is expected to perform well on both categories without overfitting.

\begin{table}[h!]
\begin{adjustbox}{max width=0.5\textwidth}
\begin{tabular}{lcccccc}
\toprule
Method & Test set & Overall Acc. & IS Acc. & IS F1 & OOS F1 & OOS recall  \\ 
      \midrule
        K+1 Classes & IND+ID-OOS & 90.65 & 90.36 & 86.46 & 90.82 & 92.76   \\ 
        discounting & IND+ID-OOS & 86.14 & 88.04 & 84.62 & 76.25 & 80.81   \\ 
        K+1 Classes & IND+OOD-OOS & 60.28 & 89.41 & 65.07 & 46.60 & 32.69   \\ 
        discounting & IND+OOD-OOS & 78.31 & 85.72 & 76.71 & 74.36 & 71.43   \\ 
        K+1 Classes & IND+both OOS types & 74.34 & 90.36 & 72.81 & 74.08 & 64.84   \\ 
        discounting & IND+both OOS types & 84.70 & 86.43 & 79.71 & 80.63 & 79.31  \\ 
\bottomrule
\end{tabular}
\end{adjustbox}
\caption{\textbf{Performance on various test sets} We compare the discounting method against the Multiclass classification method on 3 versions of test sets: IS + ID-OOS, IS + OOD-OOS (average across the datasets with OOD-OOS test examples), IS + both types of OOS examples.
}
\label{exp_discounting_vs_kp1}
\end{table}

\par \noindent \textbf{Online Evaluation}
During real-world deployment of this algorithm, we conducted additional online evaluation by analyzing the output distribution change on real production traffic 
because chatbot designers typically rely on output confidence scores to make business decisions eg. jumping to a node in the dialog tree, handing off to human agents or asking a follow-up question. Therefore, we deployed the proposed OOS algorithm in production and monitored different statistics on 10\% of randomly selected real traffic for months before surfacing it to end-users.
We observed that
more than 85\% of live traffic queries will have a less than 10\% change in top confidence after the change in OOS algorithms (Full distribution shown in Figure \ref{fig1} in Appendix). For enterprise customers with complex dialog conditions, a new algorithm that does not disrupt the normal workflow is critical for adoption.

\par \noindent \textbf{Computational Efficiency and Scalability}
Our product has a training set size limit of 25k IS and OOS training examples each and 2k IS classes. Based on this maximum training set size setting, the maximum model size for OOS detection is less 150MB based on offline testing. Based on online testing, the 99 percentiles for training time and model size of our OOS algorithm are within 30 seconds and 70MB, respectively.


\section {Conclusion}

The paper presents a novel Out of Scope (OOS) detection component in a task-oriented dialog system. It allows the assistant to recognize user input that is not designed to be answered by the assistant and need to be handed off to a human agent. For business, a well designed Out of Scope detection system can improve customer satisfaction, user engagement, lead generation and saves cost. On one hand, business wants the assistant to hand off quickly when a user input is Out of Scope. On the other hand, unnecessary hand off could increase human intervention and reduce the value of VA. We design an OOS detection system that overcomes a multitude of real-world challenges, and deploy it in production. 
\footnote{\url{https://cloud.ibm.com/docs/assistant?topic=assistant-irrelevance-detection} \url{https://cloud.ibm.com/docs/assistant?topic=assistant-release-notes\#assistant-jun162022}} 
We list out the lessons learned and both offline and online evaluation techniques for designing a robust, efficient and scalable system for enterprise VA platform.

\section*{Limitations} \label{limit}

Extensive benchmarking for other languages is out-of-scope for this work, but we have extended the approach to many European languages in the product with similar gains in performance \cite{wang2022benchmarking}. Code switching isn't evaluated in this work, but it is commonly observed in chatbots deployed in the wild. 

We have not discussed synthetic OOS examples. Despite its demonstrated effectiveness, they need caution in real world production system from a robustness perspective: it's possible to introduce spurious correlation by generated synthetic data.

\bibliography{anthology,custom}

\begin{thebibliography}{26}
\expandafter\ifx\csname natexlab\endcsname\relax\def\natexlab#1{#1}\fi

\bibitem[{Arora et~al.(2020)Arora, Jain, Chaturvedi, and Modi}]{arora2020hint3}
Gaurav Arora, Chirag Jain, Manas Chaturvedi, and Krupal Modi. 2020.
\newblock Hint3: Raising the bar for intent detection in the wild.
\newblock \emph{arXiv preprint arXiv:2009.13833}.

\bibitem[{Casanueva et~al.(2020)Casanueva, Temcinas, Gerz, Henderson, and
  Vulic}]{Casanueva2020}
I{\~{n}}igo Casanueva, Tadas Temcinas, Daniela Gerz, Matthew Henderson, and
  Ivan Vulic. 2020.
\newblock \href {https://arxiv.org/abs/2003.04807} {Efficient intent detection
  with dual sentence encoders}.
\newblock In \emph{Proceedings of the 2nd Workshop on NLP for ConvAI - ACL
  2020}.
\newblock Data available at
  https://github.com/PolyAI-LDN/task-specific-datasets.

\bibitem[{Choi et~al.(2021)Choi, Shin, Kim, and
  Shin}]{DBLP:journals/corr/abs-2105-05601}
DongHyun Choi, Myeongcheol Shin, EungGyun Kim, and Dong~Ryeol Shin. 2021.
\newblock \href {http://arxiv.org/abs/2105.05601} {Outflip: Generating
  out-of-domain samples for unknown intent detection with natural language
  attack}.
\newblock \emph{CoRR}, abs/2105.05601.

\bibitem[{Coucke et~al.(2018)Coucke, Saade, Ball, Bluche, Caulier, Leroy,
  Doumouro, Gisselbrecht, Caltagirone, Lavril, Primet, and
  Dureau}]{DBLP:journals/corr/abs-1805-10190}
Alice Coucke, Alaa Saade, Adrien Ball, Th{\'{e}}odore Bluche, Alexandre
  Caulier, David Leroy, Cl{\'{e}}ment Doumouro, Thibault Gisselbrecht,
  Francesco Caltagirone, Thibaut Lavril, Ma{\"{e}}l Primet, and Joseph Dureau.
  2018.
\newblock \href {http://arxiv.org/abs/1805.10190} {Snips voice platform: an
  embedded spoken language understanding system for private-by-design voice
  interfaces}.
\newblock \emph{CoRR}, abs/1805.10190.

\bibitem[{Gangal et~al.(2020)Gangal, Arora, Einolghozati, and
  Gupta}]{gangal2020likelihood}
Varun Gangal, Abhinav Arora, Arash Einolghozati, and Sonal Gupta. 2020.
\newblock Likelihood ratios and generative classifiers for unsupervised
  out-of-domain detection in task oriented dialog.
\newblock In \emph{Proceedings of the AAAI Conference on Artificial
  Intelligence}, volume~34, pages 7764--7771.

\bibitem[{Hemphill et~al.(1990)Hemphill, Godfrey, and
  Doddington}]{hemphill1990atis}
Charles~T Hemphill, John~J Godfrey, and George~R Doddington. 1990.
\newblock The atis spoken language systems pilot corpus.
\newblock In \emph{Speech and Natural Language: Proceedings of a Workshop Held
  at Hidden Valley, Pennsylvania, June 24-27, 1990}.

\bibitem[{Hendrycks and Gimpel(2016)}]{DBLP:journals/corr/HendrycksG16c}
Dan Hendrycks and Kevin Gimpel. 2016.
\newblock \href {http://arxiv.org/abs/1610.02136} {A baseline for detecting
  misclassified and out-of-distribution examples in neural networks}.
\newblock \emph{CoRR}, abs/1610.02136.

\bibitem[{Kamath et~al.(2020)Kamath, Jia, and
  Liang}]{DBLP:journals/corr/abs-2006-09462}
Amita Kamath, Robin Jia, and Percy Liang. 2020.
\newblock \href {http://arxiv.org/abs/2006.09462} {Selective question answering
  under domain shift}.
\newblock \emph{CoRR}, abs/2006.09462.

\bibitem[{Larson et~al.(2019)Larson, Mahendran, Peper, Clarke, Lee, Hill,
  Kummerfeld, Leach, Laurenzano, Tang et~al.}]{larson2019evaluation}
Stefan Larson, Anish Mahendran, Joseph~J Peper, Christopher Clarke, Andrew Lee,
  Parker Hill, Jonathan~K Kummerfeld, Kevin Leach, Michael~A Laurenzano,
  Lingjia Tang, et~al. 2019.
\newblock An evaluation dataset for intent classification and out-of-scope
  prediction.
\newblock \emph{arXiv preprint arXiv:1909.02027}.

\bibitem[{Lee et~al.(2018{\natexlab{a}})Lee, Lee, Lee, and
  Shin}]{lee2018training}
Kimin Lee, Honglak Lee, Kibok Lee, and Jinwoo Shin. 2018{\natexlab{a}}.
\newblock \href {https://openreview.net/forum?id=ryiAv2xAZ} {Training
  confidence-calibrated classifiers for detecting out-of-distribution samples}.
\newblock In \emph{International Conference on Learning Representations}.

\bibitem[{Lee et~al.(2018{\natexlab{b}})Lee, Lee, Lee, and
  Shin}]{NEURIPS2018_abdeb6f5}
Kimin Lee, Kibok Lee, Honglak Lee, and Jinwoo Shin. 2018{\natexlab{b}}.
\newblock \href
  {https://proceedings.neurips.cc/paper/2018/file/abdeb6f575ac5c6676b747bca8d09cc2-Paper.pdf}
  {A simple unified framework for detecting out-of-distribution samples and
  adversarial attacks}.
\newblock In \emph{Advances in Neural Information Processing Systems},
  volume~31. Curran Associates, Inc.

\bibitem[{Lewis and Gale(1994)}]{10.1007/978-1-4471-2099-5_1}
David~D. Lewis and William~A. Gale. 1994.
\newblock A sequential algorithm for training text classifiers.
\newblock In \emph{SIGIR '94}, pages 3--12, London. Springer London.

\bibitem[{Liang et~al.(2017)Liang, Li, and
  Srikant}]{DBLP:journals/corr/LiangLS17}
Shiyu Liang, Yixuan Li, and R.~Srikant. 2017.
\newblock \href {http://arxiv.org/abs/1706.02690} {Principled detection of
  out-of-distribution examples in neural networks}.
\newblock \emph{CoRR}, abs/1706.02690.

\bibitem[{Lin and Xu(2019)}]{DBLP:journals/corr/abs-1906-00434}
Ting{-}En Lin and Hua Xu. 2019.
\newblock \href {http://arxiv.org/abs/1906.00434} {Deep unknown intent
  detection with margin loss}.
\newblock \emph{CoRR}, abs/1906.00434.

\bibitem[{Liu et~al.(2019)Liu, Eshghi, Swietojanski, and
  Rieser}]{liu2019benchmarking}
Xingkun Liu, Arash Eshghi, Pawel Swietojanski, and Verena Rieser. 2019.
\newblock Benchmarking natural language understanding services for building
  conversational agents.
\newblock \emph{arXiv preprint arXiv:1903.05566}.

\bibitem[{Qi et~al.(2020)Qi, Pan, Sood, Shah, Kunc, and
  Potdar}]{DBLP:journals/corr/abs-2012-03929}
Haode Qi, Lin Pan, Atin Sood, Abhishek Shah, Ladislav Kunc, and Saloni Potdar.
  2020.
\newblock \href {http://arxiv.org/abs/2012.03929} {Benchmarking intent
  detection for task-oriented dialog systems}.
\newblock \emph{CoRR}, abs/2012.03929.

\bibitem[{Ryu et~al.(2018)Ryu, Koo, Yu, and Lee}]{ryu-etal-2018-domain}
Seonghan Ryu, Sangjun Koo, Hwanjo Yu, and Gary~Geunbae Lee. 2018.
\newblock \href {https://doi.org/10.18653/v1/D18-1077} {Out-of-domain detection
  based on generative adversarial network}.
\newblock In \emph{Proceedings of the 2018 Conference on Empirical Methods in
  Natural Language Processing}, pages 714--718, Brussels, Belgium. Association
  for Computational Linguistics.

\bibitem[{Shen et~al.(2021)Shen, Hsu, Ray, and
  Jin}]{DBLP:journals/corr/abs-2106-14464}
Yilin Shen, Yen{-}Chang Hsu, Avik Ray, and Hongxia Jin. 2021.
\newblock \href {http://arxiv.org/abs/2106.14464} {Enhancing the generalization
  for intent classification and out-of-domain detection in {SLU}}.
\newblock \emph{CoRR}, abs/2106.14464.

\bibitem[{Tax and Duin(1999)}]{DBLP:conf/esann/TaxD99}
David M.~J. Tax and Robert P.~W. Duin. 1999.
\newblock \href
  {https://www.elen.ucl.ac.be/Proceedings/esann/esannpdf/es1999-458.pdf} {Data
  domain description using support vectors}.
\newblock In \emph{{ESANN} 1999, 7th European Symposium on Artificial Neural
  Networks, Bruges, Belgium, April 21-23, 1999, Proceedings}, pages 251--256.

\bibitem[{Wang et~al.(2022)Wang, Qian, Pan, Qi, Kunc, and
  Potdar}]{wang2022benchmarking}
Gengyu Wang, Cheng Qian, Lin Pan, Haode Qi, Ladislav Kunc, and Saloni Potdar.
  2022.
\newblock Benchmarking language-agnostic intent classification for virtual
  assistant platforms.
\newblock In \emph{Proceedings of the Workshop on Multilingual Information
  Access (MIA)}, pages 69--76.

\bibitem[{Wu et~al.(2022)Wu, He, Yan, Gao, Zeng, Zheng, Zhao, Jiang, Wu, and
  Xu}]{wu-etal-2022-revisit}
Yanan Wu, Keqing He, Yuanmeng Yan, QiXiang Gao, Zhiyuan Zeng, Fujia Zheng, Lulu
  Zhao, Huixing Jiang, Wei Wu, and Weiran Xu. 2022.
\newblock \href {https://aclanthology.org/2022.naacl-main.307} {Revisit
  overconfidence for {OOD} detection: Reassigned contrastive learning with
  adaptive class-dependent threshold}.
\newblock In \emph{Proceedings of the 2022 Conference of the North American
  Chapter of the Association for Computational Linguistics: Human Language
  Technologies}, pages 4165--4179, Seattle, United States. Association for
  Computational Linguistics.

\bibitem[{Yilmaz and Toraman(2022)}]{DBLP:conf/naacl/YilmazT22}
Eyup~Halit Yilmaz and Cagri Toraman. 2022.
\newblock \href {https://aclanthology.org/2022.naacl-main.152} {{D2U:}
  distance-to-uniform learning for out-of-scope detection}.
\newblock In \emph{Proceedings of the 2022 Conference of the North American
  Chapter of the Association for Computational Linguistics: Human Language
  Technologies, {NAACL} 2022, Seattle, WA, United States, July 10-15, 2022},
  pages 2093--2108. Association for Computational Linguistics.

\bibitem[{Zeng et~al.(2021)Zeng, He, Yan, Liu, Wu, Xu, Jiang, and
  Xu}]{DBLP:journals/corr/abs-2105-14289}
Zhiyuan Zeng, Keqing He, Yuanmeng Yan, Zijun Liu, Yanan Wu, Hong Xu, Huixing
  Jiang, and Weiran Xu. 2021.
\newblock \href {http://arxiv.org/abs/2105.14289} {Modeling discriminative
  representations for out-of-domain detection with supervised contrastive
  learning}.
\newblock \emph{CoRR}, abs/2105.14289.

\bibitem[{Zhan et~al.(2021)Zhan, Liang, Liu, Fan, Wu, and
  Lam}]{DBLP:journals/corr/abs-2106-08616}
Li{-}Ming Zhan, Haowen Liang, Bo~Liu, Lu~Fan, Xiao{-}Ming Wu, and Albert Y.~S.
  Lam. 2021.
\newblock \href {http://arxiv.org/abs/2106.08616} {Out-of-scope intent
  detection with self-supervision and discriminative training}.
\newblock \emph{CoRR}, abs/2106.08616.

\bibitem[{Zhang et~al.(2022)Zhang, Hashimoto, Wan, Liu, Liu, Xiong, and
  Yu}]{zhang-etal-2022-pre-trained}
Jianguo Zhang, Kazuma Hashimoto, Yao Wan, Zhiwei Liu, Ye~Liu, Caiming Xiong,
  and Philip Yu. 2022.
\newblock \href {https://doi.org/10.18653/v1/2022.nlp4convai-1.2} {Are
  pre-trained transformers robust in intent classification? a missing
  ingredient in evaluation of out-of-scope intent detection}.
\newblock In \emph{Proceedings of the 4th Workshop on NLP for Conversational
  AI}, pages 12--20, Dublin, Ireland. Association for Computational
  Linguistics.

\bibitem[{Zhou et~al.(2022)Zhou, Liu, and Qiu}]{zhou-etal-2022-knn}
Yunhua Zhou, Peiju Liu, and Xipeng Qiu. 2022.
\newblock \href {https://doi.org/10.18653/v1/2022.acl-long.352}
  {{KNN}-contrastive learning for out-of-domain intent classification}.
\newblock In \emph{Proceedings of the 60th Annual Meeting of the Association
  for Computational Linguistics (Volume 1: Long Papers)}, pages 5129--5141,
  Dublin, Ireland. Association for Computational Linguistics.

\end{thebibliography}
\bibliographystyle{acl_natbib}

\clearpage

\appendix

\section{Appendix}

\subsection{List of IN-OOS intents}
\label{sec:appendix_dataset}
Here we list the intents for each dataset that are treated as IN-OOS intents in our benchmark.

Stackoverflow: python

SNIPS: SearchCreativeWork and SearchScreeningEvent

HAR: 
intents, hue\_lightoff, explain, remove, addcontact, wemo\_on, podcasts, createoradd, music, praise, radio, dontcare 

ROSTD: reminder/set\_reminder, reminder/cancel\_reminder, reminder/show\_reminders

HINT3 SOFMattress: SIZE\_CUSTOMIZATION, ABOUT\_SOF\_MATTRESS, LEAD\_GEN, COMPARISON, WARRANTY, DELAY\_IN\_DELIVERY

HINT3 Powerplay11: NO\_EMAIL\_CONFIRMATION, TEAM\_DEADLINE, FAKE\_TEAMS, CANNOT\_SEE\_JOINED\_CONTESTS, REFUND\_OF\_ADDED\_CASH, HOW\_TO\_PLAY, FEEDBACK, ACCOUNT\_NOT\_VERIFIED, DEDUCTED\_AMOUNT\_NOT\_RECEIVED, CRITICISM, NEW\_TEAM\_PATTERN, OFFERS\_AND\_REFERRALS 

HINT3 Curekart: EXPIRY\_DATE, CONSULT\_START, CHECK\_PINCODE, ORDER\_TAKING, INTERNATIONAL\_SHIPPING, IMMUNITY, SIDE\_EFFECT, START\_OVER, PORTAL\_ISSUE, MODES\_OF\_PAYMENTS, ORDER\_QUERY, SIGN\_UP, WORK\_FROM\_HOME

\subsection{Our OOS problem formulation is algorithm-agnostic:} 
We conducted the same experiment with another OOS algorithm: autoencoder with reconstruction loss as OOS score. The findings are similar: our OOS formulation demonstrate advantages over others. Detailed metrics are shown in Table \ref{exp_perf_ae}.

\begin{table*}[]
\begin{adjustbox}{max width=1\textwidth}
\begin{tabular}{lcccccccccc}
\toprule
Method & Overall Acc. & IS Acc. & IS F1 & OOS F1 & OOS recall & FPR90 & FPR95 & AUROC & AUPR\_IN & AUPR\_OUT  \\ 
      \midrule

        Binary & 83.25 & 85.09 & 76.68 & 81.70 & 77.16 & 28.24 & 34.78 & 91.47 & 90.37 & 89.27  \\ 
        Multiclass & 74.34 & 90.36 & 72.81 & 74.08 & 64.84 & ~ & ~ & ~ & ~ &   \\ 
        IS clf + Max & 79.17 & 85.93 & 77.99 & 70.02 & 65.15 & 32.66 & 44.45 & 87.53 & 87.15 & 80.27 \\ 
        \bf{Discounting (Our Approach)} & 84.30 & 85.85 & 77.88 & 83.17 & 79.35 & 19.65 & 25.56 & 93.18 & 92.24 & 91.63  \\ 
        
\bottomrule
\end{tabular}
\end{adjustbox}
\caption{\textbf{Performance metrics} This table compares the discounting method against the Multiclass classification method, the binary classification method, the IS classifier + max confidence on the full test sets, using autoencoder as the OOS detection algorithm.
}
\label{exp_perf_ae}
\end{table*}


\subsection{Online evaluation statistics}
\begin{figure}[h!]
\centering
\includegraphics[width=0.9\columnwidth]{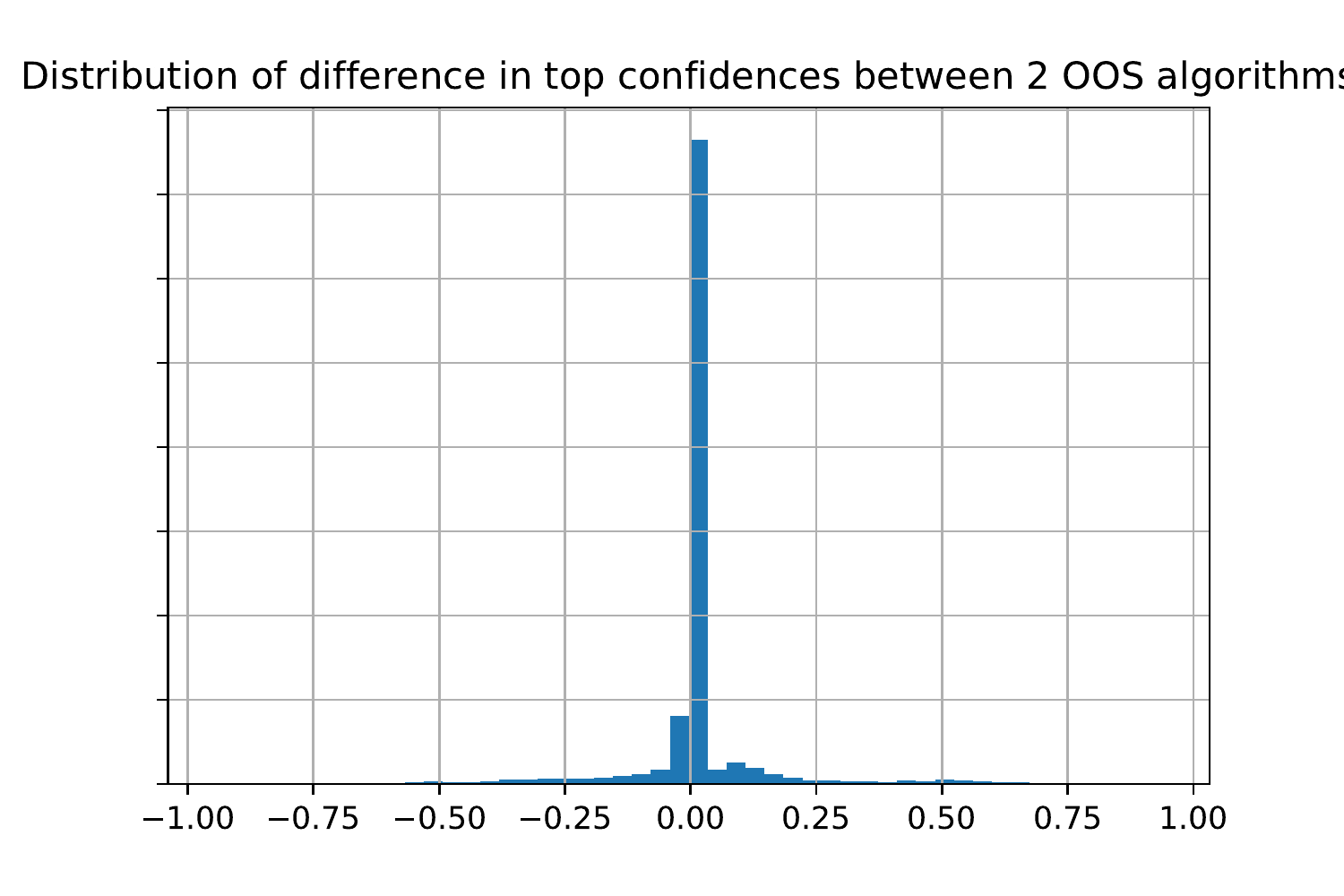} 
\caption{Distribution of differences in top confidence between the proposed OOS algorithms vs previous OOS algorithm on a percentage of live traffic}.
\label{fig1}
\end{figure}

Figure \ref{fig1} shows the full distribution of differences in top confidence between the proposed OOS algorithms vs previous OOS algorithm on a percentage of live traffic

\end{document}